# SERC: Syntactic and Semantic Sequence based Event Relation Classification

Kritika Venkatachalam
*Knowledgeable Computing and Reasoning (KRaCR) Lab*
*IIIT-Delhi*
New Delhi, India.
kritika17061@iiitd.ac.in

Raghava Mutharaju
*Knowledgeable Computing and Reasoning (KRaCR) Lab*
*IIIT-Delhi*
New Delhi, India.
raghava.mutharaju@iiitd.ac.in

Sumit Bhatia
*Media and Data Science Research Lab*
*Adobe Systems*
Noida, India.
Sumit.Bhatia@adobe.com

*Abstract*—Temporal and causal relations play an important role in determining the dependencies between events. Classifying the temporal and causal relations between events has many applications, such as generating event timelines, event summarization, textual entailment and question answering. Temporal and causal relations are closely related and influence each other. So we propose a joint model that incorporates both temporal and causal features to perform causal relation classification. We use the syntactic structure of the text for identifying temporal and causal relations between two events from the text. We extract parts-of-speech tag sequence, dependency tag sequence and word sequence from the text. We propose an LSTM based model for temporal and causal relation classification that captures the interrelations between the three encoded features. Evaluation of our model on four popular datasets yields promising results for temporal and causal relation classification.

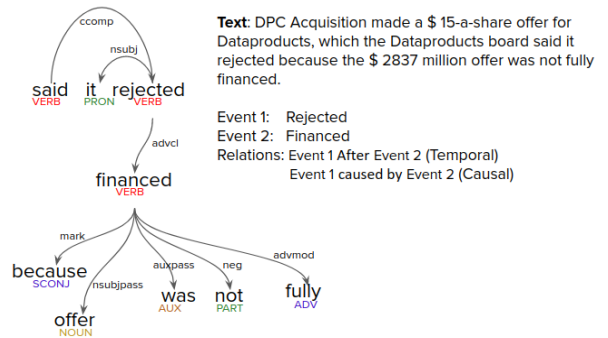

Fig. 1: An example containing two events connected through temporal and causal links and the corresponding dependency sub-tree.

## I. INTRODUCTION

Extracting temporal and causal relations between two events from a textual description has several benefits. It can help in generating event timelines [1], visualisations, text summarization [2], and in question-answering systems [3]. This is also helpful in predicting the temporal and causal links in an event knowledge graph [4]. To understand this better, consider the various events that unfolded after the recent outbreak of the COVID-19 pandemic. Some of these events were directly caused by the outbreak and hence have a causal relationship with the pandemic. Furthermore, some of the events happened after the initial steps taken to contain the spread of the virus; thereby, establishing a temporal relation between these events.

Studying the temporal relation can also help in identifying cause-effect event chains. Previous work focused on either temporal relations or causal relations but there has been minimal research on studying temporal relations as a feature that strengthens causal relation classification or vice versa. Earlier work on temporal relation classification was based on discrete features defined over lexico-syntactic and semantic structures. These features capture only explicit characteristics that indicate temporal relations and therefore fail to capture implicit characteristics present in the text that define a temporal/causal link. Some recent studies use features derived using bidirectional encoder representations from transformers [5]. Others have used sequential models with features derived along dependency paths between two events [6].

Following this, we observed that while important syntactic and semantic structures are derived along dependency paths between two event mentions in a text, the dependency path between these events does not always capture the implicit references indicating a temporal/causal relation between the events in a text. Consider the sentence, "DPC Acquisition made a $15-a-share offer for Dataproducts, which the Dataproducts board said it rejected because the $2837 million offer was not fully financed." Here, *rejected* and *financed* are the two events. These events are connected by temporal and causal relations. Figure 1 shows the dependency sub-tree for the given sentence. We can observe that the dependency path between the two events in the text completely misses the context. Thus, we use the features derived from the entire text along with the features derived along the dependency path between the two event mentions. Specifically, we use parts-of-speech (POS) tag sequence and dependency sequence for the entire text. But instead of utilising the complete word sequence, we have considered the context words that align with the dependency path for a pair of events to make the prediction specific for that event pair. The results show that including features derived from the entire text give us a better per-class performance. The contributions of this work are as follows.

- We present a sequential model for temporal and causal relation classification with stacked LSTMs that captures the inter-dependencies among the individually encoded features. The empirical results show that this has a positive effect on the overall performance.
- We propose a joint model incorporating both the temporal and causal features for causal relation classification. Our evaluation shows that utilizing both temporal and causal features of the text provide a significant performance gain in causal relation classification.

The source code and the models of our temporal and causal relation classification system, named *SERC*, are publicly available with an Apache License 2.0 at https://github.com/kracr/Temporal-Causal-Relation-Classification.

## II. RELATED WORK

There is existing work on temporal link labelling as a classification task. Mani et al. [7] designed a feature based model with a maximum entropy classifier on human annotated data that performed better than the other rule-based models for the classification of temporal relation. Chambers et al. [8] proposed a fully automatic two-stage machine learning architecture wherein the first stage knows the temporal attributes. The second stage combines them with other linguistic features for predicting temporal classes. D'Souza and Ng [9] suggested a hybrid approach by combining rule-based and learning-based systems that employ rich linguistic knowledge acquired from a variety of grammatical and discourse relations. Following this, Mirza and Tonelli [10] proved that adopting a simple feature set rather than using more complex features based on semantic role labelling and parsing resulted in better performance as compared to previous works [9].

Several works have used features along the dependency path between two event mentions in the text. Chambers et al. [11] captured the syntactic context by using the dependency path between event pairs. Laokulrat et al. [12] proposed a system of logistic regression classifiers. They used a deep syntactic parser to extract feature sets from paths between event words in phrase structure and predicate-argument structures. Choubey and Huang [6] extract features along the dependency path between two event instances in the dependency tree of the sentence. They showed that this sequential model performed better than the feature-based models like [9], [10].

When compared to the work on identify temporal relation in the text, there has been comparatively less focus on identifying causal relations. Hidey and McKeown [13] created a training set by leveraging parallel Wikipedia corpora to identify implicit markers that are variations on known causal phrases. A causal classifier was then trained using semantic features that provide contextual information. Dunietz et al. [14] use the idea of construction grammar and discussed two supervised methods for tagging causal constructions. Both the methods combine automatically produced pattern-matching rules with statistical classifiers that learn the parameters of the constructions. Do et al. [15] identified causality relations by combining distributional similarity methods and discourse connectives. They show that distributional similarity approaches improve learning event causality.

Recent works focused on both temporal and causal relations as they are closely linked. Bethard and Martin [16] annotated temporal and causal relations for 1000 event pairs and trained SVM classifiers using features acquired from WordNet and the Google N-gram corpus. Later, Mostafazadeh et al. proposed CaTeRs [17], an annotation framework for event relations in stories aiming to capture both temporal and causal aspects of the events. Moving away from a single learner, Chambers et al. [18] proposed a sieve based architecture CAEVO for ordering temporal relations. Following this, Mirza and Tonelli [19] designed CATENA, a combination of machine-learned and rule-based sieve to extract and classify temporal and causal link from English texts. Their architecture depends on events as presented in the TimeML encoding structure, that includes time stamping, ordering of events, and other temporal expressions. They also included the document creation time (DCT). Ning et al. [20] modelled this as an integer linear programming problem and presented a joint inference framework using constrained conditional models. They created a new dataset that annotates temporal and causal annotations and then showed that the joint framework, TCR (Temporal and Causal Reasoning) and the imposed constraints enhance the temporal and causal elements.

The closest related research to the one proposed by us is by Choubey and Huang [6]. They propose a sequential model for temporal relation classification between events. However, it differs from our work in the following ways.

- They extract the feature sequences aligning with the dependency path between the event mentions. In contrast, we obtain the POS and dependency sequences corresponding to the word tokens in the complete sentence. For the word sequence, we follow [6] and consider the words along the dependency path between the event mentions to make the prediction specific for the event pair under consideration.
- They focus only on the task of temporal relation classification, whereas we consider both temporal and causal relation classification. To the best of our knowledge, there has been no existing work on sequential models for causal relation classification. Furthermore, we propose a joint model exploiting the relationship among temporal and causal links to improve the prediction of causal relations.
- They propose a sequential model wherein the features are simply encoded using LSTMs. The output sequences are concatenated and passed through a dense layer for classification. Instead, our model architecture first encodes the input features. It then captures the interrelationship among these encoded features using another layer of LSTM wherein the hidden sequences of the previous layer are passed.

- They consider only intra-sentence event pairs, whereas we consider both the intra-sentence and cross-sentence event pairs.

## III. METHODOLOGY

In this section, we discuss the tasks, data preprocessing, a model architecture for temporal and causal relation classification and a joint model architecture that incorporates both temporal and causal features for causal classification.

### A. Task Description

Given two events and the corresponding text, our models classify the relationship among the events into the subtypes of temporal and causal relations. Our model is not restricted to only intra-sentence event pairs. For example, if two events in different sentences are related by temporal or causal link then the sentences are likely to have implicit/explicit references to the other event. Thus, we include cross sentence event pairs by concatenating the sentences in which the events are mentioned in the order in which they appear in the text.

Earlier work on temporal relation classification have considered only six relation types. Later, TempEval-3 [21] extended the number of target classes to 14 fine-grained temporal relations. Some of the recent work have also considered 14 classes. We evaluate our model's classification performance on both six and 14 temporal relations. Recent studies have considered three target classes for causal relation classification [19], [20]. Therefore, we follow the same for causal relation classification.

### B. Data Pre-processing

Monroe and Wang [22] and Reddy et al. [23] observed that semantic composition was related to grammatical dependency relations. Choubey and Huang [6] established that the features derived along the dependency paths between two event instances capture essential syntactic and semantic features. Along with that, they also showed that a sequential model that encoding these features contributes significantly to temporal classification. Considering that the dependency structure of natural language text captures important syntactical features, we extract syntactical and semantic feature sequences, transform them into vectors and encode the sequences using a bidirectional-LSTM (BiLSTM).

Unlike [6], for POS tag sequence and dependency tag sequence, we consider the complete sentence/text for a given sample instead of extracting along the dependency path between the events. We use the Stanford CoreNLP pipeline [24] for dependency parsing and POS tags. We then transform the input sequence into a sequence of vectors. Each token in dependency and POS tag sequence is converted into a one-hot vector. We use pre-trained GloVe embeddings [25] to transform each token in the context word sequence to a vector. Finally, these three sequences are considered as input to the model and encode the features using a BiLSTM for each.

### C. Model architecture for temporal or causal relation classification

Figure 2 illustrates the model architectures for temporal and causal relation classification. We refer to the temporal model as SERC-t and causal model as SERC-c. We experimented with different sets of nodes on the datasets from Section IV-A and achieved the best results with the following setup. The first layer of SERC-t and SERC-c consists of three parallel BiLSTMs with 64, 32 and 32 nodes. The three input sequences are context word, dependency, and POS tags and are encoded using their corresponding BiLSTM. These sequences represent the inherent syntactical and semantic framework of the text. A bidirectional-LSTM (BiLSTM) consist of two uni-directional LSTMs, which are assembled in forward and backward direction. BiLSTMs are used to capture the influence of both the patterns lying behind and ahead of the current pattern. The output of these BiLSTMs are features representing the intra syntactical relations for each sequence. These syntactical structures are interrelated. Thus, we merge the hidden layer sequences of the BiLSTMs and pass it to the next layer, a BiLSTM with 64 nodes, to further exploit the association amongst these syntactical structures.

This architecture enables the model to encode the features of the input sequences in the first layer and capture any inter-dependencies in the second layer. The second layer will learn relationships among POS tags, dependency tags and the context word sequences. After learning the inter-relationships, the output is processed through a neural network for generating classification results. The output from the second layer is passed through a dense layer with 32 nodes which is further advanced through another dense layer with the number of neurons equal to the target number of classes in the dataset. As this is a multi-class classification task, the output layer uses softmax activation to classify the temporal and causal links.

### D. Joint Model for Causal Relation Classification

In order to take advantage of the close association between temporal and causal relations, we propose a joint model, SERC-tc, that incorporates both the temporal and causal features for causal relation classification. By temporal and causal features, we mean the features extracted through temporal and causal sub-models. We explain the intuition behind using both temporal and causal features through an example shown in Figure 1. The events *rejected* and *financed* are the first and second events, respectively. Here, they are connected by both temporal and causal relations. The knowledge that the first event occurred after the second event, indicates that the first event could be a consequence of the second event. Similarly, we can conclude that a temporal relationship exists between the two events if we know that the second event caused the first event as the second event must have occurred before the first event. Previous works like [19] have explored the interaction of causal and temporal relations using multi-sieve architecture, yielding promising results. We propose a sequential model that merges both the temporal and causal features for the task of causal relation classification. To the best of our knowledge,

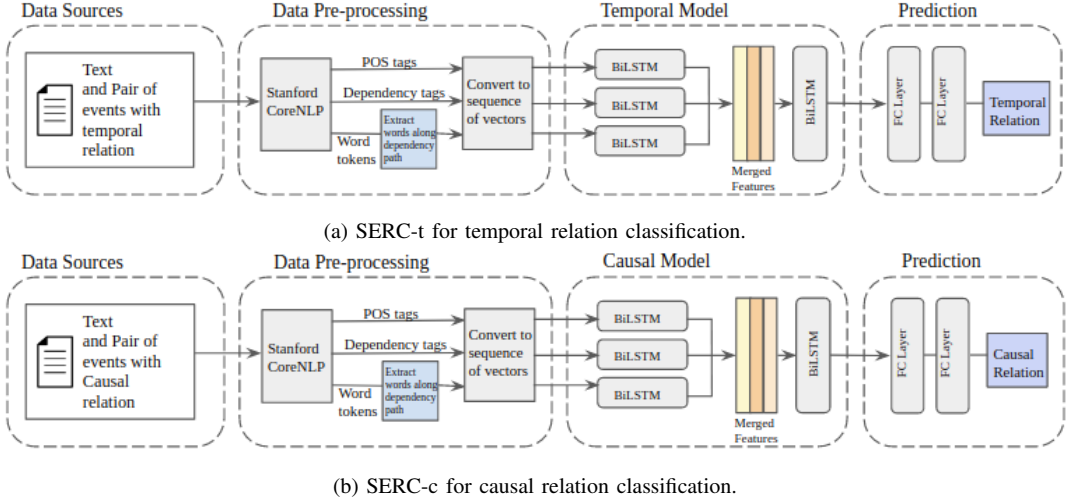

(a) SERC-t for temporal relation classification.

(b) SERC-c for causal relation classification.

Fig. 2: Model architectures to classify the temporal and causal relations between two events.

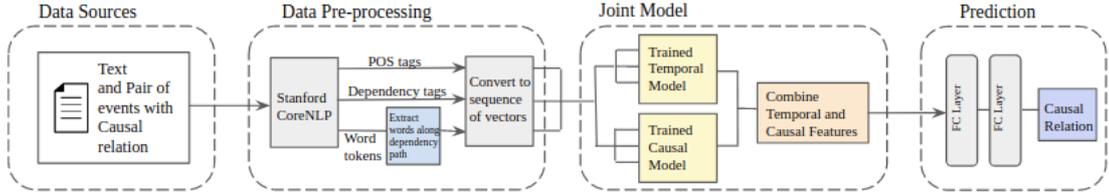

Fig. 3: SERC-tc. A joint model architecture to classify causal relations

this is the first work on using sequential model for causal relation classification.

Figure 3 illustrates the joint model architecture for causal relation classification. Using the models discussed in Section III-C, we take the output of the second layer, i.e., the stacked LSTM in SERC-t and SERC-c, to extract the temporal and causal features, respectively. The temporal and causal features are then concatenated and passed through a neural network for classification. This neural network consists of a dense layer with 32 neurons and an output layer with the number of neurons equal to the number of target classes. The output layer uses the *softmax activation function* for classification.

Furthermore, we can modify our joint model for temporal relation classification by changing the number of neurons in the output layer to the number of target classes in the temporal dataset. Causal links are much sparser than temporal links in the Causal-TimeBank corpus and the Temporal and Causal Reasoning dataset. Thus, we do not perform joint temporal relation classification due to these dataset limitations.

IV. RESULTS AND ANALYSIS

We perform four sets of evaluations to compare our model with the state-of-the-art. Two sets of experiments were for evaluating the temporal part of our model. First, we follow the TimeBank-Dense evaluation methodology, with the pre-defined train, validation and test split. We then assess our model on the TimeBank 1.2 corpus with the experimental setting from [6]. The other two sets of experiments were to evaluate the causal part of our work and compare it with two state-of-the-art works; CATENA [19], Joint Temporal and Causal Reasoning [20]. For both the evaluations, we follow the experimental setup in their respective papers.

CATENA is trained on Causal-TimeBank corpus [26] and tested on a manually annotated test set with documents from TempEval-3-platinum [21]. Ning et al. [20] released a data corpus, namely the TCR dataset [20] for joint temporal and causal reasoning with well-defined train and test splits. We focus only on event to event temporal relations. We do not compare our model's temporal part with [20] due to the unavailability of the code.

A. Datasets

We briefly describe the datasets used in the evaluation of our models.
1) **TimeBank 1.2**. The corpus [27] was developed by the Brandeis University. It contains data from 183 English news articles, annotated with events and temporal information. The corpus consists of 6,418 temporal relations and of 14 fine-grained temporal classes. These classes are after, before, begins, begun by, during, during inv, ends, ended by, iafter, ibefore, identity, includes, is included, and simultaneous [28].
2) **TimeBank-Dense**. This corpus [18] addresses the sparsity problem in the TimeML corpora [29]. It contains 12,715 temporal relations over 36 documents taken from TimeBank 1.2. For the evaluation over the TimeBank-Dense dataset, we utilise the pre-defined train, test and val splits. It comprises of a 22 document training set, a 9 document test set and a 5 document development set. We consider only the event to event temporal relations. The corpus contains six target classes; `After`, `Before`, `Simultaneous`, `Includes`, `Included In` and `Vague`.
3) **Causal-TimeBank**. This corpus [26] was created by annotating the TimeBank corpus with the temporal and causal information. It is part of the TempEval-3 English training data that has been annotated with causal data. The corpus comprises of 5,118 temporal links and 318 causal links.
4) **Temporal and Causal Reasoning Dataset**. Ning et al. [20] augmented the EventCausality dataset presented in [15] with a revised version of the dense temporal annotation scheme given in [30]. The dataset constitutes of 20 training documents and 5 testing documents with 3,400 temporal links and 172 causal links.

*B. Evaluation Results*

We evaluate our model using standard micro-average scores and accuracy in order to be consistent with the earlier studies in temporal and causal relation classification. We additionally analyse the model performance using per-class F1-scores.

**Temporal Relation Classification**. The systems we have used for comparing with SERC-t (our model) on temporal relation classification task are as follows.
- Baseline-1 is a neural network with BiLSTMs for temporal relation classification proposed by [6] with input features as a sequence of context words, parts-of-speech tag sequence and a dependency relation sequence aligning with the dependency path between two event mentions.
- Structured Joint Model [31] is a neural structured support vector machine model that extracts events and temporal relations from a given text. It performs shared representation learning and applies a global structure through ILP constraints.

| Model | F1-score |
|---|---|
| Baseline-1 | 44.2 |
| SERC-t without stacked LSTM | 46.4 |
| SERC-t with stacked LSTM | **50.0** |

TABLE I: Results on TimeBank-Dense for SERC and Baseline

We analyse the performance of our model as compared to the system proposed in [6]. Table I reports the micro-average F1-score for the baseline and our system (SERC-t). Using the complete POS tag sequence and dependency tag sequence as input features instead of just the dependency path between the events improves the classifier's performance significantly. Extracting the POS tag sequence and dependency tag sequence from the entire text captures important features indicating temporal relation. These are overlooked when we consider only the path between the two event mentions in the dependency tree. The interrelation between the encoded features captures important characteristics that indicate a temporal relation between the events. This gives us a performance gain of 5.8% over the baseline in the F1-score.

Table II[1] reports the per-class and micro average scores for all the systems on the TimeBank-Dense corpus. Structured Joint Model results on the TimeBank-Dense corpus are from Han et al. [31] as we ran into issues[2] while running the code. We follow the TimeBank-Dense evaluation methodology as stated earlier and use the pre-defined train, test and validation split. Our model performs better than Baseline-1 and the complex Structured Joint Model. It achieved the best precision, recall and F1-scores for almost all the classes except A and B. The bold values represent the best results for that particular class label. Furthermore, the per-class metrics show that our model can also identify event relations that belong to the minority classes. In contrast, both Baseline-1 and Structured Joint Model do not recognise them. The empirical results show that our system outperforms Baseline-1 by 6.8% for average precision, recall and F1-score, while it outperforms the Structured Joint Model by 3.5% in average recall and 0.6% in average F1-score.

Table III reports the per-class and micro average scores for all the systems on the TimeBank corpus that has fine-grained temporal relations. The corpus does not have pre-defined splits. Thus, we create our train, test, and validation splits with the ratio of 75%, 10% and 15%, respectively. We run both Baseline-1 and our model with the same experimental setting. Values with '-' implies that no predictions were made for that class as the data is highly imbalanced and these labels are minority classes. We can see from Table III that our model is able to identify all the classes identified by Baseline-1 and outperforms Baseline-1 for the relation types `Simultaneous`, `Is Included`, `Includes`, `Identity` and `Ended By`. The average results show that our system outperforms Baseline-1 by 3.5% for all three metrics.

Overall, for the temporal relation classification, our proposed model, SERC-t, performs well for most classes on both the datasets. Compared to the other models, our model can achieve better scores even for some of the minority classes.

**Causal Relation Classification**. The systems we have used for comparing our joint model that incorporates temporal features for causal classification task are as follows.

---

[1]- indicates that the model made no predictions for that label. The reason could be that these labels belong to minority classes.

[2]Errors with the Gurobi Optimization. The code gives an attribute error (AttributeError: Unable to retrieve attribute 'x').

| Class | Baseline-1 | | | Structured Joint Model | | | SERC-t | | |
|---|---|---|---|---|---|---|---|---|---|
| | Precision | Recall | F1 | Precision | Recall | F1 | Precision | Recall | F1 |
| A | 31.6 | 18.9 | 23.3 | **59.8** | **46.9** | **52.6** | 52.1 | 20.9 | 29.8 |
| B | - | - | - | **71.9** | **46.7** | **56.6** | 54.5 | 27.3 | 36.4 |
| S | - | - | - | - | - | - | **11.1** | **3.3** | **5.1** |
| II | - | - | - | - | - | - | **17.1** | **7.6** | **10.5** |
| I | - | - | - | - | - | - | **15.4** | **5.3** | **7.8** |
| V | 46.6 | **87.0** | 60.7 | 45.9 | 55.8 | 50.4 | **50.8** | 83.5 | **63.2** |
| Avg | 44.2 | 44.2 | 44.2 | **52.6** | 46.5 | 49.4 | 50.0 | **50.0** | **50.0** |

TABLE II: Per-class metrics for the temporal relation classification on the Time-Bank Dense corpus.

| Class | SERC-t | | | Baseline-1 | | |
|---|---|---|---|---|---|---|
| | Precision | Recall | F1 | Precision | Recall | F1 |
| Before | **35.5** | 38.6 | **37.0** | 32.0 | **42.1** | 36.4 |
| After | **51.6** | **61.1** | **55.9** | 48.4 | 56.5 | 52.1 |
| Simultaneous | **31.2** | 29.4 | **30.3** | 23.2 | 37.3 | 28.6 |
| IBefore | - | - | - | - | - | - |
| IAfter | - | - | - | - | - | - |
| Is Included | **50.0** | **36.4** | **42.1** | 25.0 | 22.7 | 23.8 |
| Includes | 18.8 | **21.4** | **20.0** | 30.0 | 10.7 | 15.8 |
| Identity | **34.1** | **46.7** | **39.4** | 31.2 | 16.7 | 21.7 |
| Begun By | - | - | - | - | - | - |
| Ended By | 100.0 | **22.2** | **36.4** | 100 | 11.1 | 20.0 |
| Begins | - | - | - | - | - | - |
| Ends | - | - | - | - | - | - |
| During | - | - | - | - | - | - |
| During Inv. | - | - | - | - | - | - |
| Avg. | **40.3** | **40.3** | **40.3** | 35.7 | 35.7 | 35.7 |

TABLE III: Per-class metrics for the temporal relation classification on the TimeBank corpus.

- Baseline-1. We modify the sequential model proposed by [6] for temporal relation classification by changing the output layer size to three classes for causal relation classification.
- CATENA. A system comprising two multi-sieve modules for temporal and causal relation classification [19]. The temporal information from the temporal model is fed into the causal classifier.
- Joint TCR model. A joint inference framework presented by [20] for temporal and causal reasoning using constrained conditional models (CCMs).

| Model | Precision | Recall | F1 |
|---|---|---|---|
| CATENA | 73.7 | 53.8 | 62.2 |
| Baseline-1 | 73.0 | 73.0 | 73.0 |
| SERC-c | 84.6 | 84.6 | 84.6 |
| SERC-tc | **92.3** | **92.3** | **92.3** |

TABLE IV: Causal relation classification results on TempEval-3 test set (manually annotated with causal links by Mirza and Tonelli [19])

| Text | Baseline 1 | SERC-c | SERC-tc |
|---|---|---|---|
| Heavy snow is causing disruption to transport across the UK, with heavy **rainfall** bringing **flooding** to the south-west of England. | causes | caused-by | caused-by |
| The **call**, which happened as President Barack Obama wrapped up his first presidential visit to Israel, was an unexpected outcome from a Mideast **trip** that seemed to yield few concrete steps. | causes | causes | caused-by |

TABLE V: A few examples from causally annotated TempEval-3 test set and class predicted by Baseline, SERC-c and SERC-tc. The events appear in bold in the text.

Table IV summarises the scores for all the systems on the TempEval-3 test set manually annotated with causal links by Mirza et al. [19]. The results of CATENA on the test set are from Mirza and Tonelli [19] which is trained on Causal-TimeBank. Our model outperforms CATENA [19] by 0.6% for temporal relation classification between events. We evaluate Baseline-1 and our model with the same experimental settings as [19] and same train, test sets for causal relation classification. We assess the performance of our joint model by incorporating both temporal and causal features. Our model outperforms Baseline-1 and CATENA by 11.6% and 22.4% in F1-score, respectively. We observe that using both temporal and causal information in our joint models improves the performance by 7.7%. This indicates that the temporal and causal relations are related, and their association helps identify causal relations significantly.

We further analyse the performance of our models on causal prediction using a few examples. Table V reports a few examples and the class identified by Baseline-1, SERC-c and SERC-tc. In both the examples, the first event is *caused by* the second event. Baseline-1 fails to predict the correct target class in the first example, but both SERC-c and SERC-tc correctly predict the causal type. We can infer that capturing the interrelations between the encoded feature sequences is crucial to predict causal relations. Baseline-1 and SERC-c fail to identify the causal link from the second event to the first event in the second example. In comparison, SERC-tc, our joint model, identifies the causal link correctly, which indicates that including temporal features is beneficial for causal relation classification.

| Model | Causal only | Causal and Temporal |
|---|---|---|
| Baseline I | 71.9 | - |
| Joint TCR Model [20] | 70 | 77.3 |
| SERC-c and SERC-tc | **73.8** (c) | **80** (tc) |

TABLE VI: Accuracy scores of our system and baselines on the TCR dataset for causal relation classification.

Table VI lists accuracy scores for all the systems on the TCR dataset. We evaluate Baseline-1 and our model with the same experimental settings as [20] and use the same train, test splits for causal relation classification. We first assess our model with only causal features, later with temporal and causal features as explained in Section III-D . Our model that fuses temporal and causal features outperforms both Baseline-1 and the Joint TCR Model by 9.2% and 2.7% respectively. Without temporal features, the performance of our model drops by 6.2%. Therefore, this reaffirms our observation that the interconnection between temporal and causal relations benefits causal relation prediction.

## V. Conclusion

Identifying temporal and causal relations between two events from textual descriptions is beneficial for several applications such as generating event timelines and question-answering systems. We presented a sequential model for temporal and causal relation classification. We devised a recurrent neural network based model that extracts temporal and causal features through syntactic and lexical sequences from the text. The model architecture enables it to capture any interrelation among these sequences. Moreover, we exploit the relationship between temporal and causal links and present a joint model incorporating both temporal and causal features for causal relation classification. We evaluate our models for temporal and causal relation classification tasks and compare it with the state-of-the-art approaches. The empirical results show that our model achieves state-of-the-art performance on both temporal and causal relation classification tasks. We have made the source code and the models publicly available for reproducibility at https://github.com/kracr/Temporal-Causal-Relation-Classification. Our evaluation proves that the connection between the temporal and causal relations improves causal relation classification. However, causal relations annotated in Causal-TimeBank and TCR dataset are sparse as compared to temporal relations. Thus, evaluating the effect of causal information in temporal relation classification is challenging. We plan to investigate this further to deal with this imbalance.

As we advance, we plan to focus on the multilingual aspects of temporal and causal relation classification as there are very few multilingual datasets. Another possible extension is investigating the role of location in the temporal and causal event relations.

**Acknowledgements.** This work is partially supported by the Infosys Center for Artificial Intelligence (CAI) at IIIT-Delhi, India.